\documentclass{article}

\usepackage[preprint]{neurips_2023}

\usepackage[utf8]{inputenc} % allow utf-8 input
\usepackage[T1]{fontenc}    % use 8-bit T1 fonts
\usepackage{url}            % simple URL typesetting
\usepackage{booktabs}       % professional-quality tables
\usepackage{amsfonts}       % blackboard math symbols
\usepackage{nicefrac}       % compact symbols for 1/2, etc.
\usepackage{microtype}      % microtypography
\usepackage{float}
\usepackage{comment}
\usepackage{graphicx}
\usepackage{amsmath}
\usepackage{bbm}
\usepackage{array}
\usepackage{amssymb}
\usepackage{caption}
\usepackage{threeparttable}
\usepackage{longtable}
\usepackage[hang,flushmargin]{footmisc}
\usepackage{multirow}
\usepackage[toc,page]{appendix}
\usepackage{url} 
\usepackage{xcolor}
\usepackage{mdframed}
\usepackage{tcolorbox}
\usepackage{lmodern}
\usepackage{lipsum}
\usepackage{marvosym}
\tcbuselibrary{skins}

\newtcolorbox{modernquote}{
  enhanced,
  colback=blue!5,
  frame hidden,
  drop shadow,
  left=10pt,
  right=10pt,
  top=10pt,
  bottom=10pt,
  boxsep=0pt,
  fontupper=\sffamily,
}

\usepackage[pagebackref,breaklinks,colorlinks]{hyperref}

\title{PLLaMa: An Open-source Large Language Model for Plant Science}

\author{Xianjun Yang\textsuperscript{1}\thanks{Equal contribution.}, Junfeng Gao\textsuperscript{2}$^*$, Wenxin Xue\textsuperscript{3}, Erik Alexandersson\textsuperscript{4} \\
\tt\small{xianjunyang@ucsb.edu, jugao@lincoln.ac.uk, xuewenxin@caas.edu.cn, erik.alexandersson@slu.se}\\
\textsuperscript{1}Department of Computer Science, University of California, Santa Barbara \quad \\
\textsuperscript{2}Lincoln Centre for Autonomous Systems, University of Lincoln \quad\\
\textsuperscript{3}Chinese Academy of Agricultural Sciences \quad\\
\textsuperscript{4}Department of Plant Breeding, Swedish University of Agricultural Sciences \quad\\
}

\begin{document}

\maketitle
\begin{abstract}

Large Language Models (LLMs) have exhibited remarkable capabilities in understanding and interacting with natural language across various sectors. However, their effectiveness is limited in specialized areas requiring high accuracy, such as plant science, due to a lack of specific expertise in these fields. This paper introduces PLLaMa, an open-source language model that evolved from LLaMa-2. It's enhanced with a comprehensive database, comprising more than 1.5 million scholarly articles in plant science. This development significantly enriches PLLaMa with extensive knowledge and proficiency in plant and agricultural sciences.
Our initial tests, involving specific datasets related to plants and agriculture, show that PLLaMa substantially improves its understanding of plant science-related topics. Moreover, we have formed an international panel of professionals, including plant scientists, agricultural engineers, and plant breeders. This team plays a crucial role in verifying the accuracy of PLLaMa's responses to various academic inquiries, ensuring its effective and reliable application in the field.
To support further research and development, we have made the model's checkpoints and source codes accessible to the scientific community. These resources are available for download at \url{https://github.com/Xianjun-Yang/PLLaMa}.

\end{abstract}

\section{Introduction}

The advancements in large language model (LLM) technologies, exemplified by OpenAI's ChatGPT~\citep{chatgpt} and GPT-4~\citep{openai2023gpt4}, represent a significant stride in artificial intelligence, enhancing applications such as machine translation \citep{zhang2023prompting}, text summarization \citep{yang2023exploring}, among others. However, the detailed methodologies and architectures of ChatGPT and its variants remain largely undisclosed. The substantial API costs associated with these models pose a barrier to their widespread application in various fields.

Conversely, publicly accessible foundational models like LLaMA-2~\citep{touvron2023llama} sometimes underperform in specialized tasks, likely due to the absence of domain-specific data in their initial training, such as in the field of plant science~\citep{plantblindbess}.

Efforts to distill insights from proprietary models for instruction-based tuning have seen notable success. Models like Alpaca~\citep{alpaca} and Vicuna~\citep{vicuna2023} focus on improving interactive abilities using samples generated by machines for instruction following. Therefore, the divide between publicly accessible models and proprietary products continues to narrow, bridging the distance in their capabilities and features.

In this context, we present PLLaMa, an open-source language model developed by extending the training of LlaMa-2-7B and LlaMa-2-13B \citep{touvron2023llama} with a dataset of over 1.5 million academic papers in plant science. The resulting models exhibit enhanced performance in addressing plant science-related queries.

\begin{figure}[t]
    \centering
    \includegraphics[width=\textwidth]{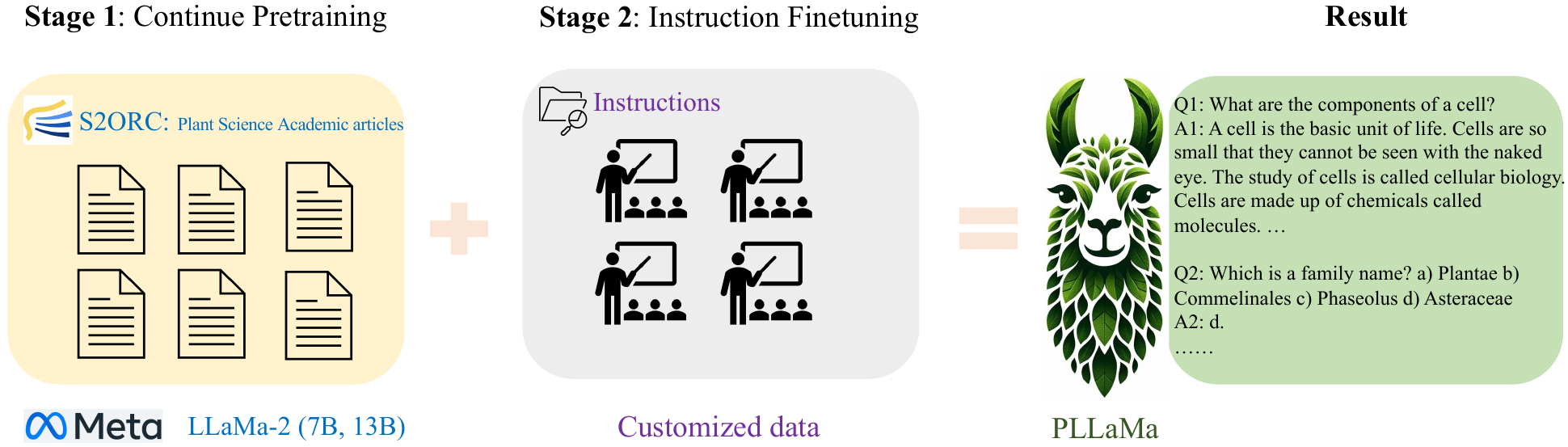}
    \caption{PLLaMa Training Pipeline.
    }
    \vspace{-10pt}
    \label{fig:teaser}
\end{figure}

The process comprises two key stages, as depicted in Figure~\ref{fig:teaser}: an initial phase of extended pretraining with abundant academic articles in plant science, followed by a phase of instruction-based fine-tuning.

\section{Related Work}

The architecture of Transformers \citep{vaswani2017attention} along with the objective of predicting the next word have greatly enhanced auto-regressive models such as GPT-2 \citep{radford2019language} and GPT-3 \citep{brown2020language}. 
Then, instruction tuning has been pivotal in enhancing the conversational capabilities of foundational models, making a standard approach for leading commercial chatbots like ChatGPT \citep{schulman2022chatgpt} and GPT-4 \citep{openai2023gpt4}.

Furthermore, the emergence of open-source foundational models including LLaMa \citep{touvron2023llama}, Falcon \citep{refinedweb}, LLaMa-2 \citep{touvron2023llama}, and OPT \citep{zhang2022opt} have provided developers with robust platforms for diverse product development across various domains, including medicine \citep{Han2023MedAlpacaA} and law \citep{Huang2023LawyerLT}.

Although NLP techniques have been widely used for various tasks, ranging from material science \citep{yang2022pcmsp, yang2023matkb, yang2024quokka}, finance \citep{wu2023bloomberggpt}, natural science \citep{xie2023darwin}, geology science \citep{Lin2023GeoGalacticaAS}, medical \citep{zhang2023alpacare, Wu2023PMCLLaMAFF} and astronomy \citep{nguyen2023astrollama},
the adaptation of large language models to plant science has lagged behind. 
To address this void, we utilize extensive databases such as S2ORC \citep{lo-wang-2020-s2orc} to extend the pretraining of language models, incorporating an increased amount of plant science information. This process demands significant computational resources. Therefore, we're providing all our training checkpoints at no cost to the research community.

\begin{table}[!ht]
\centering

\begin{tabular}{|p{0.3\linewidth}|p{0.6\linewidth}|}
\hline
\textbf{Question} & \textbf{Answer} \\
\hline
How can cells endocytose against the turgor pressure? & 
- The cell membrane is flexible. When the cell engulfs a vesicle, parts of the cell membrane that are not in contact with the vesicle relax, making room for the vesicle.
- The cell membrane is permeable to water. As the vesicle is engulfed, water from the cell flows into the vesicle, reducing its pressure.
- The vesicle is smaller than the cell membrane pores. The vesicle sits in the pores like a marble in a cola straw. Since the straw can easily deform, it collapses a little around the marble. In the same way, the cell membrane collapses a little around the vesicle, reducing the pressure. \\
\hline
How does cytokinesis separate the cytoplasmic volume? & 
Cytokinesis is the process by which the cell divides its contents and forms two daughter cells. In animal cells, it occurs through a process called phragmoplasmic reticulum (P-body) expansion, which results in the formation of a new cell plate. The P-body is a network of invaginations that forms around the plasma membrane during the G1 phase of the cell cycle. As the cell enters S phase, the P-body expands and elongates, pushing against the plasma membrane and creating a new cell plate. This new cell plate eventually separates the cell into two daughter cells. \\
\hline
How do sperm cells reach the egg? & 
- The sperm cells are transported in the male's semen.
- During intercourse, some of the semen is expelled from the vagina, and some remains inside it.
- The cervix produces mucus, which is thick and similar to glue. It acts as a barrier that prevents sperm from entering the womb.
- The mucus slowly becomes thinner over the course of several days, becoming thinner and more like water.
- When the time is right for ovulation, the mucus becomes very thin, and some sperm cells are able to swim out of the semen and navigate through the cervix.
- The sperm cells continue to swim through the female reproductive system, and one of them reaches the egg, which is released from the ovary at around the same time. \\
\hline
How can the organ of a living organism bend and exert forces? & 
Suppose you have a piece of plastic that is very flexible. You can bend it and it will return to its original shape. Now take that piece of plastic and stretch it. It will become more rigid. This is because stretching the plastic makes the molecules that make up the plastic move apart. The same thing happens in your muscles. When you relax, your muscles contract and molecules in your muscles move closer together. This makes your muscles more rigid. When you flex your muscles, those same molecules move apart, making your muscles more flexible. \\
\hline
\end{tabular}

\caption{Questions and Answers generated by PLLaMa-13B-Chat}\label{tab: qa_res}
\end{table}

\section{Experiment}

\textbf{Plant Science Journals} We utilize the Internet to collect all journal names related to Plant Science, resulting in a total of 750 distinct journals. Then, we collect all the articles in the S2ORC \citep{lo-wang-2020-s2orc} dataset by either satisfying including 'Plant' in the article publisher or the publisher falling in the 750 journals. As a result, we obtained $1,676,389$ plant science-related articles.

\begin{table}[h!]
\centering
\scalebox{0.85}{
\begin{tabular}{|p{0.18\linewidth}|p{0.77\linewidth}|p{0.05\linewidth}|}
\hline
\small{ \textbf{Question} } &\small{  \textbf{Choices} } &\small{  \textbf{Ours.} } \\
\hline
1) Which method(s) a breeder typically uses in order to generate new genetic variants in his/her plant & 
A)	different irrigation treatments on selected plants
B)	crossing of selected plants
C)	tissue culture
D)	gene editing using Crispr/Cas9

Correct answers: B &
C \\
\hline
 2)	A molecular marker in a typical plant breeding context … & 
A)	detects insertion/deletion types of DNA polymorphisms
B)	separates different tissue types in a plant, such as roots from leaves
C)	measures the macronutrient and micronutrient content in plant tissue
D)	detects single nucleotide polymorphisms in the DNA 

Correct answers: 
A, D &
A
 \\
\hline
3)	Which of the “Mendelian Rules” is NOT universally correct.  & 
A)	the rule of uniformity of the F1 after crossing  pure line parents
B)	the rule of independent segregation of two genes
C)	the rule that genes are inherited in two versions: one from the pollen and one from the egg cell parent
D)	the rule that genes do not get lost from parents to offspring even if they may remain invisible (recessive) 

Correct answers: 
B & C
 \\
\hline
 4)	The domestication of the first crop plants in human history happened & 
 A)	in Eastern Asia, Yangtze valley, 25.000 years ago
B)	in West Africa, Nile valley 600.000 years ago
C)	in West Asia, Euphrates and Tigris valley, 10.000 years ago
D)	in Southern Europe, Po valley, 3.000 years ago

Correct answers: 
 C & C
 \\
\hline
 5)	The Green Revolution was & 
A)	A series of technical innovations particularly in Asia since the 1960s, such as the introduction of short straw cultivars
B)	The re-discovery of the Mendelian Rules in the year 1900, leading to a boost in plant breeding
C)	A series of innovations in plant breeding in America, promoted through the North American land grant universities between 1920 and 1950 
D)	A series of public protests and revolutionary events in North Africa and West Asia, triggered by rising food prices from 2010 to 2015 

Correct answers: 
A & C
 \\
\hline
6)	The Breeding Value of an individual is defined as follows  & 
A)	The average performance of this individual, compared to a population mean
B)	The general performance of this individual, compared to the best currently available cultivars
C)	The average performance of the progeny of this individual, compared to a population mean
D)	The average performance of the progeny of this individual, compared to the best existing cultivars 

Correct answers: 
C & C
 \\
\hline
7)	What is described with the term Genetic Erosion?  & 
 A)	Climate change induced heavy rain and floods that swamp seed, soil and plants into the ocean
B)	Loss of genetic diversity in crops due to replacement of landraces by modern cultivars
C)	Loss of genetic diversity in wild plants due to permanent change of the ecosystem, e.g. deforestation
D)	UV-light induced mutations that mutate and permanently change important genes in a plant

Correct answers: 
B, C & B, C
 \\
\hline
8)	What is a typical measure for ex-situ conservation of genetic resources  & 
A)	Collecting plants or seeds in nature and storing them in a gene-bank
B)	Collecting plants or seeds in nature and growing them in a botanical garden
C)	Collecting plants or seeds from breeders and storing them in a gene-bank
D)	Protecting a natural habitat where wild species are growing 

Correct answers: 
A, B, C & A, B, C
 \\
\hline
9)	The definition of Heterosis is as follows:   & 
A)	A trait (e.g. grain yield) in a hybrid is higher than in the better parent
B)	A trait (e.g. grain yield) in a hybrid is higher than in the lower parent
C)	A trait (e.g. grain yield) in a hybrid is higher than the mid-parent value
D)	A trait (e.g. grain yield) in a hybrid is equal to the better parent 

Correct answers: 
C & C
 \\
\hline
 10)	The Heritabiltiy Coefficient is a measure for & 
A)	The total number of genes in a plant species
B)	The relative contribution of the genetic variance compared to the total phenotypic variance for a trait
C)	The proportion of expressed genes in a plant species
D)	The relative yield advantage of new cultivars compared to old cultivars 

Correct answers: 
B  & D
 \\
 \hline
\end{tabular}
}
\caption{Questions and our answer prediction generated by PLLaMa-13B-Chat}\label{tab: 10_quiz}
\end{table}

\textbf{Corpus}: The Llama-2 model underwent its initial training using a comprehensive compilation of text derived from the web. This initial training phase was focused on equipping the chatbot with a basic grasp of common sense. It covered a wide range of subjects, terms, and conceptual structures that are commonly found in human knowledge. However, the model has not been fine-tuned for specific domains.
So we collect the plant science domain corpus for continued pretraining.
Based on the $1,676,389$ plant science articles, we set the chunk window to be $5,120$, resulting in $2,123,154$ text pieces.  
We also mix the material corpus with $10\%$ (typically, $93,051$ text) of the general RedPajama-Data-1T-Sample dataset \footnote{https://huggingface.co/datasets/togethercomputer/RedPajama-Data-1T-Sample}, to prevent catastrophic forgetting of general knowledge. So the total pretraining text pieces is $2,278,433$. Notice that the chunk window is larger than our following model training max length of $1024$ tokens, so there is still abundant room for additional training to fully utilize the corpus. We leave this for future work due to the resource limit.

\begin{figure}[t]
    \centering
    \includegraphics[width=\textwidth]{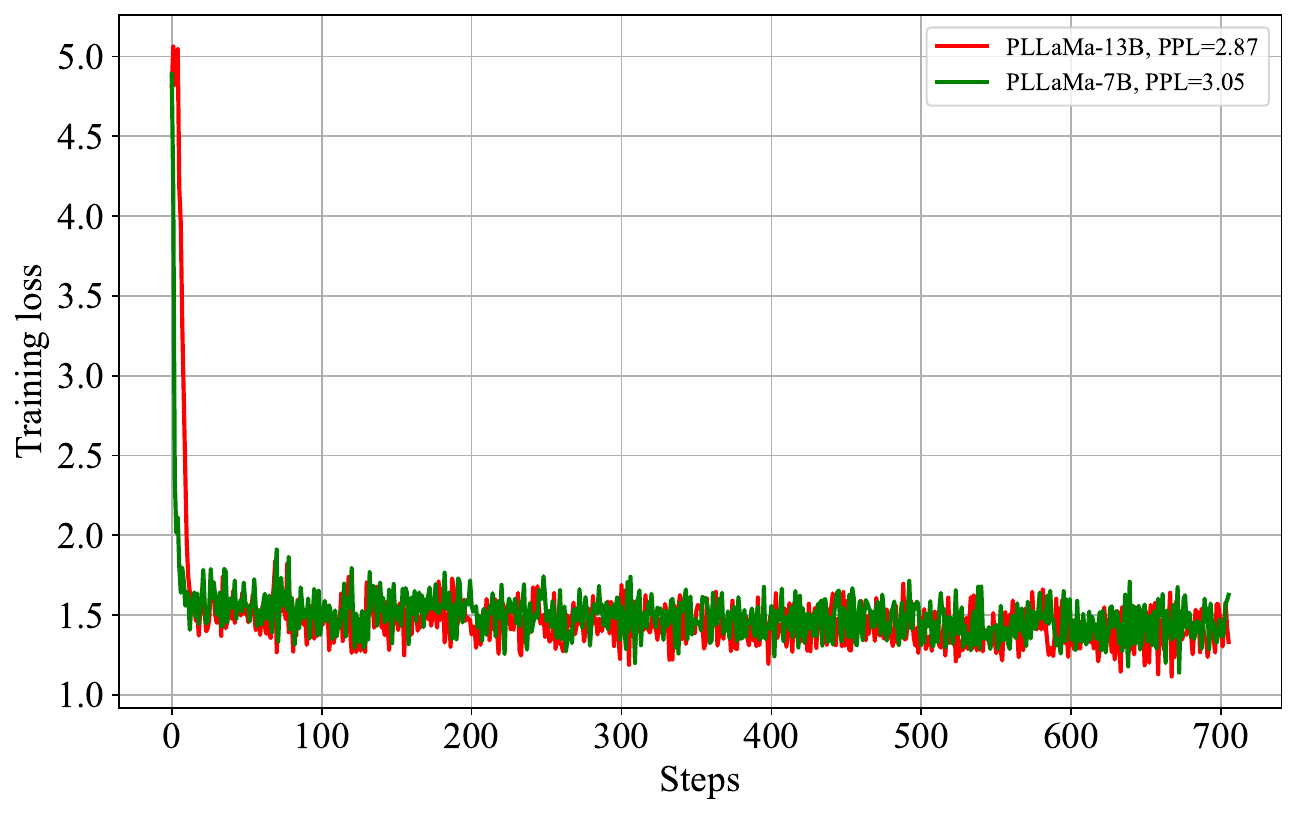}
    \caption{Pretraining loss curve.
    }
    \vspace{-10pt}
    \label{fig:training_loss_2}
\end{figure}

\textbf{Experimental Configuration}:\\
For pretraining, we utilized eight A100 80G GPUs. Our training involved a maximum token length of 1024, and we incorporated bf16 and flash-attention \citep{dao2022flashattention} to enhance training efficiency. Additionally, zero-stage-3 from DeepSpeed \footnote{https://github.com/microsoft/DeepSpeed} was employed. Each device had a batch size of 2, with gradient accumulation after every 200 steps. The initial learning rate was 2e-5. We opted for no direct weight decay and a cosine lr scheduler. Warm-up was crucial to avoid model collapse, leading us to choose a warm-up ratio of 0.3. The Fully Sharded Data Parallel (FSDP) pipeline from Hugging Face \footnote{https://huggingface.co/docs/accelerate/usage\_guides/fsdp} was used. Both the 7B and 13B models underwent a single epoch of pretraining to prevent overfitting. The training duration for these models on 8 A100 GPUs was approximately 26 hours for the 7B model and 57 hours for the 13B model. The pretraining loss is plotted in Figure \ref{fig:training_loss_2}.

\begin{figure}[t]
    \centering
    \includegraphics[width=\textwidth]{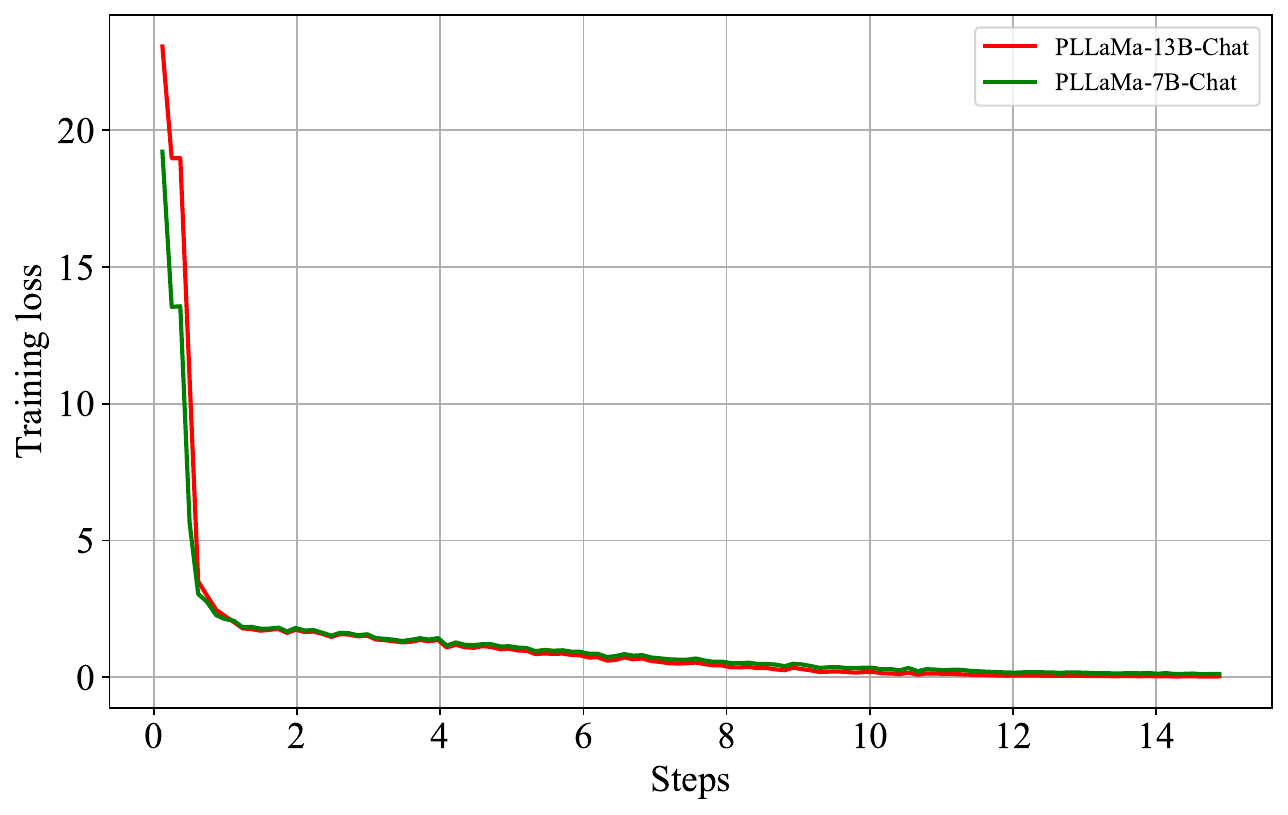}
    \caption{Instruction tuning loss curve.
    }
    \vspace{-10pt}
    \label{fig:instruction_loss}
\end{figure}

Instruction Tuning Process: Post-pretraining, the model was subjected to instruction tuning to enhance its proficiency in understanding and responding to specific instructions or queries. We incorporated 1030 instructions from the LIMA \citep{zhou2023lima} training set and also developed additional customized instructions focusing on plant science, with the intention of expanding this in future work.
For instruction tuning, we employed four A100 80G GPUs. We set the number of epochs at 15 for both the 7B and 13B models, with a learning rate of $1e-4$, a warm-up ratio of $0.3$, and a maximum token length of 1024. The batch size per device was two, and we chose a gradient accumulation step interval of 16. The lr scheduler was a cosine type. The FSDP pipeline with bf16 precision was also utilized. The instruction-tuning process took roughly 1.3 hours for the 7B model and 2.7 hours for the 13B model, based on the 1030 instructions. The instruction tuning loss is plotted in Figure \ref{fig:instruction_loss}.

\section{Benchmark}
\label{sec:benchmark}
 
We first evaluate it on the 10 held-out plant science quiz and get around $60\%$ accuracy on the multi-choice questions. The full quiz can be seen in Table \ref{tab: 10_quiz}. This shows the usefulness of our model.

\section{Zero-shot Case Study}
In this section, we show the model answers to zero-shot questions on PLLaMa-13B-Chat in Table \ref{tab: qa_res}. Our internal team of global experts confirmed that those answers are fairly accurate and useful. A more detailed evaluation is left for the next version.

\section{ Conclusion and Future Work}

In this report, we make efforts to strengthen the fundamental language models with domain-specific knowledge of plant science. We utilize the plant science journal names for filtering to get relevant academic articles, resulting in over 1.5 million articles. Then we perform continued pertaining on this plant science corpus based on LLaMa-2 7B and 13B base models. This process involves significant computation so we make the trained checkpoints freely available to the community. We also perform instruction tuning to empower the model with dialogue ability.

In the future, we plan to release a more comprehensive instruction-tuning dataset to embrace more diverse domain-specific instructions, aiming at a more specified model for plant science. We will also make a more thorough evaluation soon.

\bibliographystyle{plainnat}
\bibliography{egbib}
%%%%%%%%%%%%%%%%%%%%%%%%%%%%%%%%%%%%%%%%%%%%%%%%%%%%%%%%%%%%

\clearpage
\section{Appendix}

We first list all the 750 plant science journal names in the follwoing: 

['Rice',
 'FLORA',
 'GRANA',
 'NOVON',
 'TAXON',
 'ALGAE',
 'Foods',
 'BLUMEA',
 'Botany',
 'Fottea',
 'PLANTA',
 'Webbia',
 'Arctoa',
 'Lilloa',
 'MAYDICA',
 'PRESLIA',
 'RHODORA',
 'SYDOWIA',
 'TELOPEA',
 'Vegetos',
 'Lazaroa',
 'Rheedea',
 'Nuytsia',
 'BioRisk',
 'BIOLOGIA',
 'BOTHALIA',
 'Bradleya',
 'CALDASIA',
 'CASTANEA',
 'GENETICA',
 'Gorteria',
 'Wulfenia',
 'HERZOGIA',
 'TUEXENIA',
 'MycoKeys',
 'NeoBiota',
 'ADANSONIA',
 'BRITTONIA',
 'CANDOLLEA',
 'CYTOLOGIA',
 'EUPHYTICA',
 'MYCOTAXON',
 'Phytotaxa',
 'PhytoKeys',
 'Hacquetia',
 'Karstenia',
 'Rostaniha',
 'BRYOLOGIST',
 'HASELTONIA',
 'MYCORRHIZA',
 'PHYCOLOGIA',
 'PLANT CELL',
 'AoB Plants',
 'Mycosphere',
 'Plant Root',
 'Darwiniana',
 'Lindbergia',
 'Check List',
 'Plant Gene',
 'Bonplandia',
 'Kitaibelia',
 'AGROCIENCIA',
 'BMC BIOLOGY',
 'CYTOTHERAPY',
 'Plant Omics',
 'PROTOPLASMA',
 'Willdenowia',
 'Rhizosphere',
 'Rodriguesia',
 'Agriculture',
 'Reinwardtia',
 'AEROBIOLOGIA',
 'IAWA JOURNAL',
 'WEED SCIENCE',
 'Plant Genome',
 'KEW BULLETIN',
 'Crop Journal',
 'Rice Science',
 'Plants-Basel',
 'EFSA Journal',
 'Plant Direct',
 'Plant Stress',
 'Bio-protocol',
 'Dendrobiology',
 'HARMFUL ALGAE',
 'NOVA HEDWIGIA',
 'PLANT BIOLOGY',
 'PLANT DISEASE',
 'PLANT ECOLOGY',
 'PLANT JOURNAL',
 'Plant Methods',
 'PLANT SCIENCE',
 'WEED RESEARCH',
 'Alpine Botany',
 'Organogenesis',
 'Nature Plants',
 'Horticulturae',
 'Turczaninowia',
 'Lankesteriana',
 'Austrobaileya',
 'Biodiversitas',
 'Italus Hortus',
 'EPPO Bulletin',
 'AQUATIC BOTANY',
 'Fungal Ecology',
 'PHYTOCHEMISTRY',
 'PHYTOPATHOLOGY',
 'PLANT AND SOIL',
 'PLANT ARCHIVES',
 'PLANT BREEDING',
 'Planta Daninha',
 'Planta Daninha',
 'ACTA AMAZONICA',
 'Botany Letters',
 'Herba Polonica',
 'Coffee Science',
 'Czech Mycology',
 'Field Mycology',
 'Biotechnologia',
 'Legume Science',
 'Bioelectricity',
 'BOTANICA MARINA',
 'ECONOMIC BOTANY',
 'GAYANA BOTANICA',
 'Legume Research',
 'Molecular Plant',
 'NEW PHYTOLOGIST',
 'PHOTOSYNTHETICA',
 'PHYTOCOENOLOGIA',
 'PHYTOPARASITICA',
 'PLANT PATHOLOGY',
 'TRANSPLANTATION',
 'TREE PHYSIOLOGY',
 'WEED TECHNOLOGY',
 'GAYANA BOTANICA',
 'NEW PHYTOLOGIST',
 'Plant Diversity',
 'Plant Sociology',
 'Genes and Cells',
 'Acta Mycologica',
 'Journal of Nuts',
 'Natura Croatica',
 'Forest Research',
 'Transplantology',
 'BOTANICAL REVIEW',
 'BREEDING SCIENCE',
 'PLANT BIOSYSTEMS',
 'PLANT PHYSIOLOGY',
 'TROPICAL ECOLOGY',
 'Journal of Fungi',
 'Botanica Serbica',
 'Italian Botanist',
 'Biota Colombiana',
 'in silico Plants',
 'British Wildlife',
 'Oil Crop Science',
 'Studies in Fungi',
 'Transplantologia',
 'BMC PLANT BIOLOGY',
 'Botanical Studies',
 'DENDROCHRONOLOGIA',
 'FOLIA GEOBOTANICA',
 'Genetika-Belgrade',
 'GRASSLAND SCIENCE',
 'SYSTEMATIC BOTANY',
 'GRASSLAND SCIENCE',
 'Botanical Studies',
 'Botanical Studies',
 'Agriculture-Basel',
 'Flora Montiberica',
 'ACTA AGROBOTANICA',
 'Botanica Pacifica',
 'Applied Phycology',
 'BIOLOGIA PLANTARUM',
 'JOURNAL OF ECOLOGY',
 'MOLECULAR BREEDING',
 'PLANT CELL REPORTS',
 'Botanical Sciences',
 'Botanical Sciences',
 'Plant Reproduction',
 'Annali di Botanica',
 'Fermentation-Basel',
 'Forest and Society',
 'Feddes Repertorium',
 'Flora Mediterranea',
 'ALLELOPATHY JOURNAL',
 'JOURNAL OF BRYOLOGY',
 'JOURNAL OF FORESTRY',
 'Plant Biotechnology',
 'XENOTRANSPLANTATION',
 'Phytobiomes Journal',
 'Plant Science Today',
 'Waldokologie Online',
 'New Disease Reports',
 'CELL TRANSPLANTATION',
 'JOURNAL OF PHYCOLOGY',
 'Horticulture Journal',
 'Mediterranean Botany',
 'Collectanea Botanica',
 'Contributii Botanice',
 'Journal of Phytology',
 'Plant Communications',
 'Plants People Planet',
 'AMERICAN FERN JOURNAL',
 'CRYPTOGAMIE ALGOLOGIE',
 'CRYPTOGAMIE BRYOLOGIE',
 'LIVER TRANSPLANTATION',
 'PHYCOLOGICAL RESEARCH',
 'PHYSIOLOGIA PLANTARUM',
 'PLANT SPECIES BIOLOGY',
 'SEED SCIENCE RESEARCH',
 'Transplant Immunology',
 'Horticulture Research',
 'Life Science Alliance',
 'AGRICULTURAL RESEARCH',
 'Indian Phytopathology',
 'Botanica Complutensis',
 'Botanicheskii Zhurnal',
 'Journal of Stem Cells',
 'Current Plant Biology',
 'Plant Health Progress',
 'Botanikai Kozlemenyek',
 "Rastitel'nost' Rossii",
 'Frontiers in Agronomy',
 'Plant Phenome Journal',
 'ACTA BOTANICA CROATICA',
 'Acta Botanica Mexicana',
 'CALIFORNIA AGRICULTURE',
 'Crop \& Pasture Science',
 'PHYTOCHEMICAL ANALYSIS',
 'Phytochemistry Letters',
 'PHYTOCHEMISTRY REVIEWS',
 'Crop \& Pasture Science',
 'Tropical Plant Biology',
 'Transplantation Direct',
 'Modern Phytomorphology',
 'Summa Phytopathologica',
 'Opuscula Philolichenum',
 'Ecologica Montenegrina',
 'Acta Botanica Brasilica',
 'CELL AND TISSUE BANKING',
 'FOREST PRODUCTS JOURNAL',
 'Horticultura Brasileira',
 'JOURNAL OF ETHNOBIOLOGY',
 'JOURNAL OF PEST SCIENCE',
 'PHOTOSYNTHESIS RESEARCH',
 'PLANT GROWTH REGULATION',
 'PLANT MOLECULAR BIOLOGY',
 'TRENDS IN PLANT SCIENCE',
 'Transplantation Reviews',
 'JOURNAL OF ETHNOBIOLOGY',
 'Clinical Kidney Journal',
 'Acta Botanica Hungarica',
 'Transplantation Reports',
 'Ornamental Horticulture',
 'Acta Biologica Sibirica',
 'Phytopathology Research',
 'Ratarstvo i Povrtarstvo',
 'ANNALES BOTANICI FENNICI',
 'CLINICAL TRANSPLANTATION',
 'Comparative Cytogenetics',
 'IHERINGIA SERIE BOTANICA',
 'JOURNAL OF PLANT BIOLOGY',
 'Journal of Plant Ecology',
 'NATURAL PRODUCT RESEARCH',
 'NORDIC JOURNAL OF BOTANY',
 'TRANSPLANT INTERNATIONAL',
 'Tropical Plant Pathology',
 'Journal of Plant Ecology',
 'PLANT PROTECTION SCIENCE',
 'Tropical Plant Pathology',
 'Archives Animal Breeding',
 'Harvard Papers in Botany',
 'Journal of Biopesticides',
 'Plant Physiology Reports',
 'Natural History Sciences',
 'Agriculture and Forestry',
 'Advances in Weed Science',
 'Annals of Transplantation',
 'JOURNAL OF PHYTOPATHOLOGY',
 'JOURNAL OF PLANT RESEARCH',
 'MICROBES AND ENVIRONMENTS',
 'MOLECULAR PLANT PATHOLOGY',
 'PEDIATRIC TRANSPLANTATION',
 'PLANT AND CELL PHYSIOLOGY',
 'Plant Ecology \& Diversity',
 'TURKISH JOURNAL OF BOTANY',
 'Annals of Forest Research',
 'Journal of Berry Research',
 'Biodiversity Data Journal',
 'Acta Botanica Venezuelica',
 'Acta Prataculturae Sinica',
 'Acta Horticulturae Sinica',
 'Renal Replacement Therapy',
 'Research in Plant Disease',
 'Iranian Journal of Botany',
 'AMERICAN JOURNAL OF BOTANY',
 'JOURNAL OF PLANT PATHOLOGY',
 'PAKISTAN JOURNAL OF BOTANY',
 'PLANT CELL AND ENVIRONMENT',
 'Frontiers in Plant Science',
 'Plant Signaling \& Behavior',
 'Egyptian Journal of Botany',
 'Journal of Cotton Research',
 'Thaiszia Journal of Botany',
 'Journal fur Kulturpflanzen',
 'Journal of Crop Protection',
 'Korean Journal of Mycology',
 'Quantitative Plant Biology',
 'ACTA PHYSIOLOGIAE PLANTARUM',
 'BONE MARROW TRANSPLANTATION',
 'JOURNAL OF PLANT PHYSIOLOGY',
 'PLANT BIOTECHNOLOGY JOURNAL',
 'Plant Biotechnology Reports',
 'Plant Ecology and Evolution',
 'Progress in Transplantation',
 'Records of Natural Products',
 'Revista Fitotecnia Mexicana',
 'SEED SCIENCE AND TECHNOLOGY',
 'TRANSPLANTATION PROCEEDINGS',
 'WOOD SCIENCE AND TECHNOLOGY',
 'Brazilian Journal of Botany',
 'Brazilian Journal of Botany',
 'Horticultural Plant Journal',
 'Journal of Crop Improvement',
 'Edinburgh Journal of Botany',
 'Folia Cryptogamica Estonica',
 'Mikologiya I Fitopatologiya',
 'Biopesticides International',
 'Italian Journal of Mycology',
 'Notulae Scientia Biologicae',
 'Annual Plant Reviews Online',
 'AUSTRALASIAN PLANT PATHOLOGY',
 'AUSTRALIAN JOURNAL OF BOTANY',
 'AUSTRALIAN SYSTEMATIC BOTANY',
 'BANGLADESH JOURNAL OF BOTANY',
 'EUROPEAN JOURNAL OF AGRONOMY',
 'JOURNAL OF APPLIED PHYCOLOGY',
 'PHYTOPATHOLOGIA MEDITERRANEA',
 'SYSTEMATICS AND BIODIVERSITY',
 'TREES-STRUCTURE AND FUNCTION',
 'Indian Journal of Nematology',
 "Khimiya Rastitel'nogo Syr'ya",
 'Plant and Fungal Systematics',
 'Progress in Plant Protection',
 'Historia Naturalis Bulgarica',
 'EUROPEAN JOURNAL OF PHYCOLOGY',
 'JOURNAL OF VEGETATION SCIENCE',
 'NEW ZEALAND JOURNAL OF BOTANY',
 'Transplant Infectious Disease',
 'Journal of Plant Interactions',
 'Ethnobiology and Conservation',
 'Thai Forest Bulletin (Botany)',
 'Acta Agronomica Sinica(China)',
 'PlantEnvironment Interactions',
 'Turkish Journal of Nephrology',
 'Advances in Botanical Research',
 'Annual Review of Plant Biology',
 'CHEMISTRY OF NATURAL COMPOUNDS',
 'JOURNAL OF APICULTURAL SCIENCE',
 'JOURNAL OF EXPERIMENTAL BOTANY',
 'Natural Product Communications',
 'Applications in Plant Sciences',
 'International Journal on Algae',
 'Journal of Plant Biotechnology',
 'Cell and Organ Transplantology',
 'Metabarcoding and Metagenomics',
 'Integrative Organismal Biology',
 'Journal of Pollination Ecology',
 'Annual Review of Phytopathology',
 'INDIAN JOURNAL OF BIOTECHNOLOGY',
 'PLANT SYSTEMATICS AND EVOLUTION',
 'SOUTH AFRICAN JOURNAL OF BOTANY',
 'ANNALS OF AGRICULTURAL SCIENCES',
 'Banats Journal of Biotechnology',
 'Journal of Applied Horticulture',
 'Chinese Journal of Rice Science',
 'Current Transplantation Reports',
 'Asian Journal of Plant Sciences',
 'Revista de Biologia Neotropical',
 'CURRENT OPINION IN PLANT BIOLOGY',
 'ISRAEL JOURNAL OF PLANT SCIENCES',
 'PLANT MOLECULAR BIOLOGY REPORTER',
 'Propagation of Ornamental Plants',
 'SOIL SCIENCE AND PLANT NUTRITION',
 'Australasian Plant Disease Notes',
 'Environmental Control in Biology',
 'Chinese Journal of Plant Ecology',
 'Arab Journal of Plant Protection',
 'Plant Breeding and Biotechnology',
 'Eurasian Journal of Soil Science',
 'KOREAN JOURNAL OF PLANT TAXONOMY',
 'Journal of Tropical Life Science',
 'CANADIAN JOURNAL OF PLANT SCIENCE',
 'PLANT PHYSIOLOGY AND BIOCHEMISTRY',
 'Hellenic Plant Protection Journal',
 'Journal of Horticultural Research',
 'Indian Journal of Transplantation',
 'Journal of Horticultural Sciences',
 'Journal of New Zealand Grasslands',
 'Korean Journal of Transplantation',
 'Heart Vessels and Transplantation',
 'Australian Journal of Crop Science',
 'CRITICAL REVIEWS IN PLANT SCIENCES',
 'JOURNAL OF GENERAL PLANT PATHOLOGY',
 'JOURNAL OF PLANT GROWTH REGULATION',
 'Revista Brasileira de Fruticultura',
 'Journal of Integrative Agriculture',
 'Annales, Series Historia Naturalis',
 'Pakistan Journal of Phytopathology',
 'Current protocols in plant biology',
 'Chinese Journal of Eco-Agriculture',
 'Crop Forage \& Turfgrass Management',
 'AMERICAN JOURNAL OF TRANSPLANTATION',
 'CANADIAN JOURNAL OF PLANT PATHOLOGY',
 'EUROPEAN JOURNAL OF FOREST RESEARCH',
 'EUROPEAN JOURNAL OF PLANT PATHOLOGY',
 'INTEGRATIVE AND COMPARATIVE BIOLOGY',
 'NEPHROLOGY DIALYSIS TRANSPLANTATION',
 'RUSSIAN JOURNAL OF PLANT PHYSIOLOGY',
 'Acta Phytotaxonomica et Geobotanica',
 'Natural Products and Bioprospecting',
 'Linye Kexue/Scientia Silvae Sinicae',
 'Iranian Journal of Plant Physiology',
 'Revista de la Facultad de Agronomia',
 'ACTA SOCIETATIS BOTANICORUM POLONIAE',
 'ANALES DEL JARDIN BOTANICO DE MADRID',
 'Bangladesh Journal of Plant Taxonomy',
 'GENETIC RESOURCES AND CROP EVOLUTION',
 'JOURNAL OF AGRONOMY AND CROP SCIENCE',
 'Journal of Animal and Plant Sciences',
 'Journal of Integrative Plant Biology',
 'Journal of Systematics and Evolution',
 'VEGETATION HISTORY AND ARCHAEOBOTANY',
 'Journal of Systematics and Evolution',
 'JOURNAL OF PLANT PROTECTION RESEARCH',
 'Cellular Therapy and Transplantation',
 'Neotropical Biology and Conservation',
 'Revista Chapingo, Serie Horticultura',
 'Journal of Asia-Pacific Biodiversity',
 'Electronic Journal of Plant Breeding',
 'Revista de Ciencias Agroveterinarias',
 'Natural Volatiles and Essential Oils',
 'Revista del Jardin Botanico Nacional',
 'Transplantation and Cellular Therapy',
 'Agrosystems Geosciences  Environment',
 'ENVIRONMENTAL AND EXPERIMENTAL BOTANY',
 'Invasive Plant Science and Management',
 'Environmental Technology \& Innovation',
 'Journal of Integrated Pest Management',
 'Asian Pacific Journal of Reproduction',
 'Crop, Forage and Turfgrass Management',
 'Novosti Sistematiki Nizshikh Rastenii',
 'Revista Cubana de Plantas Medicinales',
 'Ethnobotany Research and Applications',
 'ACS Agricultural Science \& Technology',
 'International Journal of Fruit Science',
 'International Journal of Plant Biology',
 'ANNALS OF THE MISSOURI BOTANICAL GARDEN',
 'INTERNATIONAL JOURNAL OF PLANT SCIENCES',
 'JOURNAL OF THE TORREY BOTANICAL SOCIETY',
 'NEW ZEALAND JOURNAL OF FORESTRY SCIENCE',
 'Tropical Grasslands-Forrajes Tropicales',
 'Scientific Papers-Series B-Horticulture',
 'Transplant Research and Risk Management',
 'International Journal of Phytopathology',
 'South African Journal of Plant and Soil',
 'Indian Journal of Agricultural Research',
 'BOTANICAL JOURNAL OF THE LINNEAN SOCIETY',
 'Current Opinion in Organ Transplantation',
 'Journal of Plant Diseases and Protection',
 'NJAS-WAGENINGEN JOURNAL OF LIFE SCIENCES',
 'Zeitschrift fur Arznei- \& Gewurzpflanzen',
 'Nephrology, Dialysis and Transplantation',
 'Revista de Investigaciones Agropecuarias',
 'Korean Journal of Medicinal Crop Science',
 'Vegetation Classification and Survey VCS',
 'Urban Agriculture  Regional Food Systems',
 'Experimental and Clinical Transplantation',
 'INTERNATIONAL JOURNAL OF PHYTOREMEDIATION',
 'International Journal of Plant Production',
 'JOURNAL OF HEART AND LUNG TRANSPLANTATION',
 'PAKISTAN JOURNAL OF AGRICULTURAL SCIENCES',
 'Journal of Animal and Plant Sciences-JAPS',
 'Journal of Crop Science and Biotechnology',
 'Ege Universitesi Ziraat Fakultesi Dergisi',
 'Acta Scientiarum Polonorum-Hortorum Cultus',
 'Horticulture Environment and Biotechnology',
 'JOURNAL OF APPLIED BOTANY AND FOOD QUALITY',
 'Horticulture Environment and Biotechnology',
 'JOURNAL OF APPLIED BOTANY AND FOOD QUALITY',
 'PHYSIOLOGY AND MOLECULAR BIOLOGY OF PLANTS',
 'Current Research in Translational Medicine',
 'International Journal of Vegetable Science',
 'Journal of Plant Resources and Environment',
 'International Journal of Forestry Research',
 'Journal of Plant Nutrition and Fertilizers',
 'USDA Forest Service - Research Note PNW-RN',
 'ACTA BIOLOGICA CRACOVIENSIA SERIES BOTANICA',
 'BIOLOGY OF BLOOD AND MARROW TRANSPLANTATION',
 'JOURNAL OF APPLIED BOTANY AND FOOD QUALITY',
 'PHYSIOLOGY AND MOLECULAR BIOLOGY OF PLANTS',
 'Current Research in Translational Medicine',
 'International Journal of Vegetable Science',
 'Journal of Plant Resources and Environment',
 'International Journal of Forestry Research',
 'Journal of Plant Nutrition and Fertilizers',
 'USDA Forest Service - Research Note PNW-RN',
 'ACTA BIOLOGICA CRACOVIENSIA SERIES BOTANICA',
 'BIOLOGY OF BLOOD AND MARROW TRANSPLANTATION',
 'JOURNAL OF APPLIED BOTANY AND FOOD QUALITY',
 'PHYSIOLOGY AND MOLECULAR BIOLOGY OF PLANTS',
 'Current Research in Translational Medicine',
 'International Journal of Vegetable Science',
 'Journal of Plant Resources and Environment',
 'International Journal of Forestry Research',
 'Journal of Plant Nutrition and Fertilizers',
 'USDA Forest Service - Research Note PNW-RN',
 'ACTA BIOLOGICA CRACOVIENSIA SERIES BOTANICA',
 'BIOLOGY OF BLOOD AND MARROW TRANSPLANTATION',
 'JOURNAL OF APPLIED BOTANY AND FOOD QUALITY',
 'PHYSIOLOGY AND MOLECULAR BIOLOGY OF PLANTS',
 'Current Research in Translational Medicine',
 'International Journal of Vegetable Science',
 'Journal of Plant Resources and Environment',
 'International Journal of Forestry Research',
 'Journal of Plant Nutrition and Fertilizers',
 'USDA Forest Service - Research Note PNW-RN',
 'ACTA BIOLOGICA CRACOVIENSIA SERIES BOTANICA',
 'BIOLOGY OF BLOOD AND MARROW TRANSPLANTATION',
 'JOURNAL OF APPLIED BOTANY AND FOOD QUALITY',
 'PHYSIOLOGY AND MOLECULAR BIOLOGY OF PLANTS',
 'Current Research in Translational Medicine',
 'International Journal of Vegetable Science',
 'Journal of Plant Resources and Environment',
 'International Journal of Forestry Research',
 'Journal of Plant Nutrition and Fertilizers',
 'USDA Forest Service - Research Note PNW-RN',
 'ACTA BIOLOGICA CRACOVIENSIA SERIES BOTANICA',
 'BIOLOGY OF BLOOD AND MARROW TRANSPLANTATION',
 'JOURNAL OF APPLIED BOTANY AND FOOD QUALITY',
 'PHYSIOLOGY AND MOLECULAR BIOLOGY OF PLANTS',
 'Current Research in Translational Medicine',
 'International Journal of Vegetable Science',
 'Journal of Plant Resources and Environment',
 'International Journal of Forestry Research',
 'Journal of Plant Nutrition and Fertilizers',
 'USDA Forest Service - Research Note PNW-RN',
 'ACTA BIOLOGICA CRACOVIENSIA SERIES BOTANICA',
 'BIOLOGY OF BLOOD AND MARROW TRANSPLANTATION',
 'JOURNAL OF APPLIED BOTANY AND FOOD QUALITY',
 'PHYSIOLOGY AND MOLECULAR BIOLOGY OF PLANTS',
 'Current Research in Translational Medicine',
 'International Journal of Vegetable Science',
 'Journal of Plant Resources and Environment',
 'International Journal of Forestry Research',
 'Journal of Plant Nutrition and Fertilizers',
 'USDA Forest Service - Research Note PNW-RN',
 'ACTA BIOLOGICA CRACOVIENSIA SERIES BOTANICA',
 'BIOLOGY OF BLOOD AND MARROW TRANSPLANTATION',
 'JOURNAL OF APPLIED BOTANY AND FOOD QUALITY',
 'PHYSIOLOGY AND MOLECULAR BIOLOGY OF PLANTS',
 'Current Research in Translational Medicine',
 'International Journal of Vegetable Science',
 'Journal of Plant Resources and Environment',
 'International Journal of Forestry Research',
 'Journal of Plant Nutrition and Fertilizers',
 'USDA Forest Service - Research Note PNW-RN',
 'ACTA BIOLOGICA CRACOVIENSIA SERIES BOTANICA',
 'BIOLOGY OF BLOOD AND MARROW TRANSPLANTATION',
 'the plant cell',
 'plant physiology',
 'journal of experimental botany',
 'plant journal',
 'plant, cell \& environment',
 'frontiers in plant science',
 'american journal of botany',
 'plant molecular biology',
 'plant science',
 'botanical journal of the linnean society',
 'trends in plant science',
 'plant biotechnology journal',
 'plant methods',
 'plant direct',
 'planta',
 'functional plant biology',
 'plant pathology',
 'weed science',
 'plant ecology',
 'plant physiology and biochemistry',
 'journal of plant growth regulation',
 'plant breeding',
 'plant disease',
 'plant genome',
 'plant physiology reports',
 'botany',
 'journal of systematics and evolution',
 'journal of plant biology',
 'plant reproduction',
 'plant species biology',
 'agriculture, ecosystems \& environment',
 'agricultural systems',
 'field crops research',
 'agricultural and forest meteorology',
 'agricultural water management',
 'sustainability',
 'agricultural research \& technology',
 'organic agriculture',
 'journal of sustainable agriculture',
 'international journal of agricultural sustainability',
 'agricultural economics',
 'agricultural finance review',
 'agricultural history review',
 'agriculture and human values',
 'archives of agronomy and soil science',
 'computers and electronics in agriculture',
 'precision agriculture',
 'journal of agricultural and applied economics',
 'journal of agricultural and resource economics',
 'journal of agricultural science',
 'journal of agribusiness in developing and emerging economies',
 'journal of agricultural and environmental ethics',
 'agricultural science',
 'agronomy',
 'agricultural research',
 'journal of integrative agriculture',
 'agronomy journal',
 'european journal of agronomy',
 'theoretical and applied genetics',
 'crop science',
 'soil science society of america journal',
 'journal of agronomy and crop science',
 'plant and soil',
 'agronomy for sustainable development',
 'soil \& tillage research',
 'journal of plant nutrition and soil science',
 'european journal of soil science',
 'soil research',
 'soil use and management',
 'journal of soil and water conservation',
 'soil science and plant nutrition',
 'soil biology and biochemistry',
 'geoderma',
 'pedosphere',
 'journal of agronomy',
 'journal of crop improvement',
 'journal of soil science and plant nutrition',
 'agronomy research',
 'soil and sediment contamination',
 'soil science',
 'journal of plant nutrition',
 'land degradation \& development',
 'soil science annual',
 'eurasian soil science',
 'biosystems engineering',
 'agricultural engineering international: cigr journal',
 'journal of agricultural science and technology',
 'transactions of the asabe',
 'agricultural information worldwide',
 'biosystems \& agriculture engineering',
 'journal of agricultural engineering',
 'agricultural engineering',
 'international agrophysics',
 'international journal of agricultural and biological engineering',
 'agricultural mechanization in asia, africa and latin america',
 'journal of agricultural machinery and bioresources engineering',
 'journal of the korean society for agricultural machinery',
 'ecology letters',
 'trends in ecology \& evolution',
 'ecology',
 'ecological applications',
 'ecological monographs',
 'journal of ecology',
 'journal of applied ecology',
 'functional ecology',
 'ecology and evolution',
 'ecological economics',
 'ecological indicators',
 'ecological modelling',
 'ecological engineering',
 'ecological complexity',
 'ecological informatics',
 'global ecology and biogeography',
 'ecosystems',
 'oecologia',
 'restoration ecology',
 'landscape ecology',
 'marine ecology progress series',
 'freshwater biology',
 'aquatic ecology',
 'forest ecology and management',
 'journal of vegetation science',
 'ecological research',
 'journal of animal ecology',
 'urban ecosystems',
 'biodiversity and conservation',
 'environmental ecology',
 'plant biology',
 'plant cell reports',
 'plant cell tissue and organ culture',
 'plant systematics and evolution',
 'plant molecular biology reporter',
 'agricultural sciences',
 'agricultural and environmental letters',
 'agricultural history',
 'agricultural and forest entomology',
 'journal of agricultural education and extension',
 'agricultural science research journal',
 'journal of agricultural extension',
 'agricultural communications',
 'agricultural administration',
 'agricultural sciences in china',
 'indian journal of agricultural sciences',
 'asian journal of agriculture and development',
 'russian agricultural sciences',
 'agricultural science and technology',
 'african journal of agricultural research',
 'australian journal of crop science',
 'advances in agronomy',
 'communications in soil science and plant analysis',
 'the journal of agricultural science and technology',
 'crop, forage \& turfgrass management',
 'applied turfgrass science',
 'journal of cotton research',
 'paddy and water environment',
 'soil and tillage research',
 'arid land research and management',
 'soil and environment',
 'soil \& water research',
 'soil horizons',
 'soil technology',
 'journal of agricultural engineering research',
 'agricultural engineering journal',
 'journal of agricultural machinery',
 'plants, people and planet',
 'phytopathology',
 'phytobiome',
 'molecular plant-microbe interactions',
 'new phytologist',
 'molecular plant',
 'pest management science',
 'biological control',
 'journal of pest science',
 'physiologia plantarum',
 'molecular plant pathology',
 'european journal of plant pathology',
 'annual review of phytopathology',
 'horticulture research',
 'journal of integrative plant biology',
 'environmental and experimental botany']

\end{document}